\def\year{2022}\relax

\pdfoutput=1
\documentclass[letterpaper]{article}
\usepackage{aaai22}
\usepackage{times}
\usepackage{helvet}
\usepackage{courier}
\usepackage[hyphens]{url}
\usepackage{graphicx}
\usepackage{natbib}
\usepackage{caption}
\DeclareCaptionStyle{ruled}{labelfont=normalfont,labelsep=colon,strut=off}
\usepackage{algorithm}
\usepackage{algorithmic}

\usepackage{newfloat}
\usepackage{listings}
\floatstyle{ruled}
\newfloat{listing}{tb}{lst}{}
\floatname{listing}{Listing}

\title{A Framework for Institutional Risk Identification using Knowledge Graphs and Automated News Profiling}
\author{
    Mahmoud Mahfouz\textsuperscript{\rm 1}, Armineh Nourbakhsh\textsuperscript{\rm 1}, Sameena Shah\textsuperscript{\rm 1}
    }
\affiliations{
    \textsuperscript{\rm 1}J.P. Morgan Artificial Intelligence Research\\
    383 Madison Avenue\\
    New York, 10017, USA\\
    \{mahmoud.a.mahfouz, armineh.nourbakhsh, sameena.shah\}@jpmchase.com
}

\begin{document}

\maketitle

\begin{abstract}
    Organizations around the world face an array of risks impacting their operations globally. It is imperative to have a robust risk identification process to detect and evaluate the impact of potential risks before they materialize. Given the nature of the task and the current requirements of deep subject matter expertise, most organizations utilize a heavily manual process. In our work, we develop an automated system that (a) continuously monitors global news, (b) is able to autonomously identify and characterize risks, (c) is able to determine the proximity of reaching triggers to determine the distance from the manifestation of the risk impact and (d) identifies organization's operational areas that may be most impacted by the risk. Other contributions also include: (a) a knowledge graph representation of risks and (b) relevant news matching to risks identified by the organization utilizing a neural embedding model to match the textual description of a given risk with multi-lingual news.
\end{abstract}

\section{Introduction \& Related Work} \label{sec:introduction}

    Global institutions are exposed to various types of risks, ranging from market risks related to the institution's core functions \cite{cfibank} \cite{boebank} to operational, compliance, cyber-security, geopolitical and reputational risks \cite{matellio}. Risks in these areas are inherently hard to identify and quantify. Risk mitigation is also extremely challenging, which is why the runway provided by its identification and quantification is crucial. Unfortunately, the lack of proper risk assessment has led to the demise of several organizations once the risk manifested \cite{de2012international}. In our work, we present a system for risk identification utilizing knowledge graphs for representing risk areas and a neural embedding model \cite{reimers2019sentence} for multi-lingual news matching tailored towards financial institutions.

    The formal definition of a risk and the study of methods for risk assessment and mitigation has a long history of academic research \cite{henley1981reliability}, \cite{covello1985risk}, \cite{rechard1999historical}, \cite{bedford2001probabilistic}, \cite{thompson2005interdisciplinary} and \cite{zio2009reliability}. Similarly, the use of natural language processing and knowledge graphs for news recommendation has been studied extensively and is used widely in practice. The majority of the work has focused on developing news recommendation systems tailored towards users preferences. \cite{wang2020discovery} describes a news recommendation system employed by a major financial ratings agency utilizing a neural embedding model developed by \cite{peters2018deep} for news contextual embeddings representation. Other approaches to news recommendations include \cite{ijntema2010ontology} which utilize externally developed ontologies to find news, collaborative filtering \cite{lu2015content} and graph embeddings \cite{ren2019financial}.
    
\section{System Architecture} \label{sec:approach}
    
\begin{figure}[!htb]
        \centering
        \includegraphics[width=0.35\textwidth]{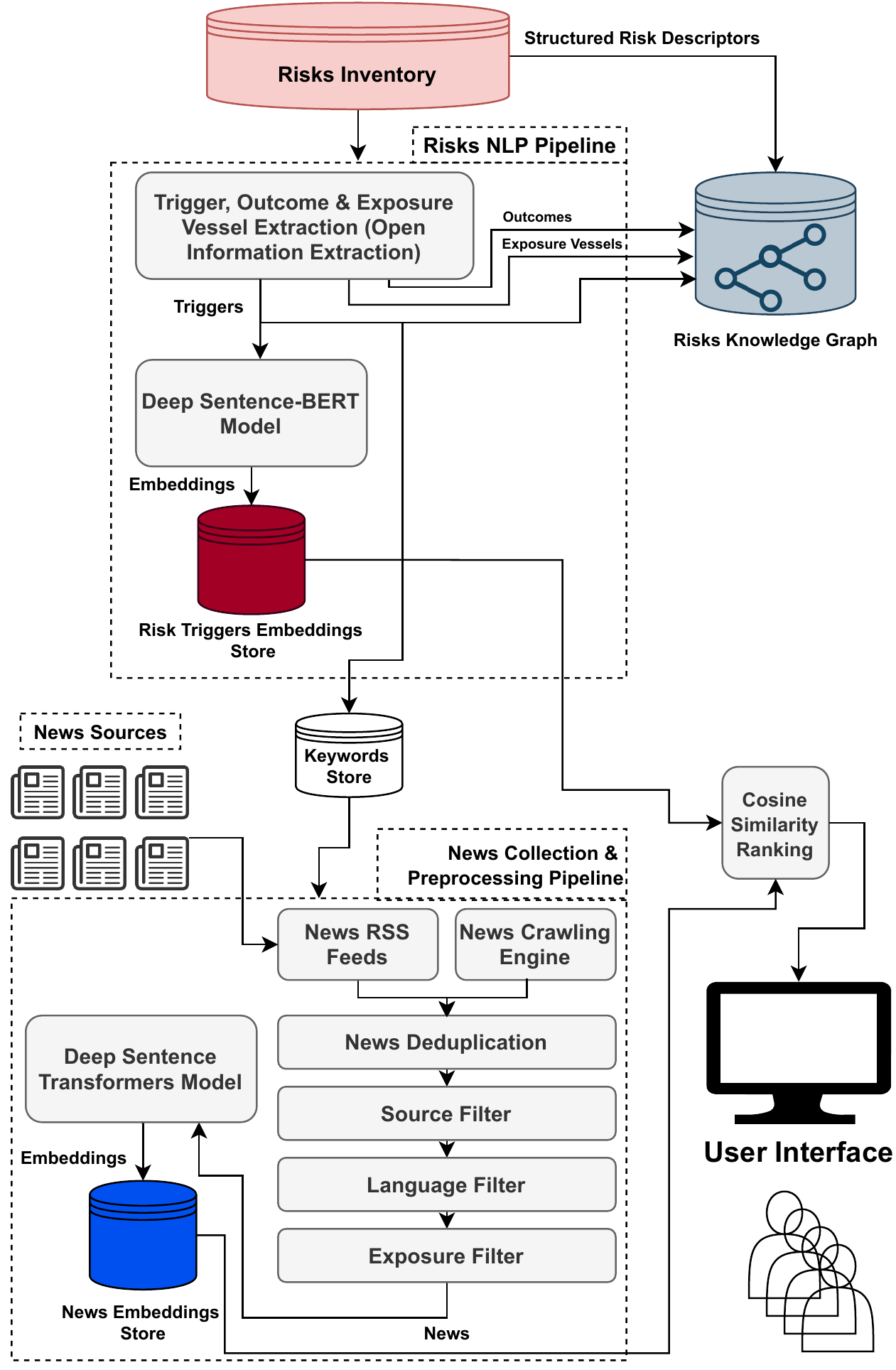}
        \caption{System architecture}
        \label{fig:sa}
    \end{figure}

    The system proposed (Figure \ref{fig:sa}) consists of four main components starting with a given set of risks identified by domain experts and producing a list of relevant news for each risk. The input to the system is a repository of risk descriptions provided by the domain experts. The first component extracts a set of relevant entities from the textual description of the risk. The second component uses these entities to construct a knowledge graph. The third component searches for a set of keywords related to the risk and parses them from multiple news sources. The fourth component is a neural network model used to rank news events using contextual embeddings generated for headlines as well as the risk descriptors.
    
    In order to demonstrate the end-to-end workflow of our system, we created a set of artificial risks shown in table \ref{table:risks} which a financial institution would face inspired from the risk types defined in \cite{matellio}. 
    
    \begin{center}
         \begin{table}[h!]
             \begin{tabular}{||c||} 
             \hline
                 \begin{tabular}{@{}l@{}}(1) Cyber-attacks targeting the retail banking busin-\\ess causing a loss of customer data\end{tabular} \\
             \hline
                \begin{tabular}{@{}l@{}}(2) US - China trade war escalation affecting the\\ corporate and investment banking business causing\\ a decrease in revenues\end{tabular} \\
             \hline
                \begin{tabular}{@{}l@{}}(3) Employee misconduct in the investment banki-\\ng business causing a reputational damage\end{tabular} \\
             \hline
                 \begin{tabular}{@{}l@{}}(4) Technology infrastructure failure in the corpor-\\ate and investment banking business causing a rep-\\utational damage and/or monetary loss\end{tabular} \\
             \hline
            \end{tabular}
        \caption{Examples of Institutional Risks}
        \label{table:risks}
        \end{table}
        \vspace{-8mm}
    \end{center}
    
    \subsection{Risk Information Extraction} \label{sec:rie}
    
        Given a textual description of the risk, the text is decomposed into (1) \textit{trigger} (the root cause of the risk), (2) \textit{outcome} (the impact of the given risk) and (3) \textit{exposure vessel} (the entity/vessel the risk impacts).
        
\begin{figure}[!htb]
            \centering
            \includegraphics[width=0.45\textwidth]{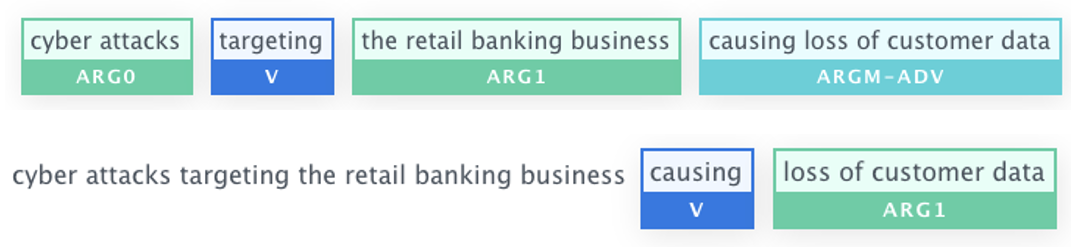}
            \caption{Risk information extraction example \protect\cite{gardner2018allennlp}.}
            \label{fig:allen_nlp_oie}
        \end{figure}

        Several approaches were tested to decompose the text into the three categories above. One of these is based on a deep bi-LSTM neural network sequence prediction model developed by \cite{stanovsky2018supervised} for supervised open information extraction. The model breaks a given sentence (in our case the risk text) into the relationships they express. In particular, the model extracts a list of propositions, each composed of a single predicate and an arbitrary number of arguments \cite{stanovsky2018supervised}. As an example, consider the risk (1) in table \ref{table:risks}. The model breaks the sentence into the components described in figure \ref{fig:allen_nlp_oie}.
        
        In this example the first argument would map to the trigger of the risk which in this case is cyber attacks. The outcome and exposure vessel typically follow the first verb in the sentence. In this example, the retail banking business is the exposure vessel and the argument after the second verb, loss of customer data, is the outcome.

    \subsection{Risks Knowledge Graph} \label{sec:rkg}
        
\begin{figure}[!htb]
                \centering
                \includegraphics[width=0.45\textwidth]{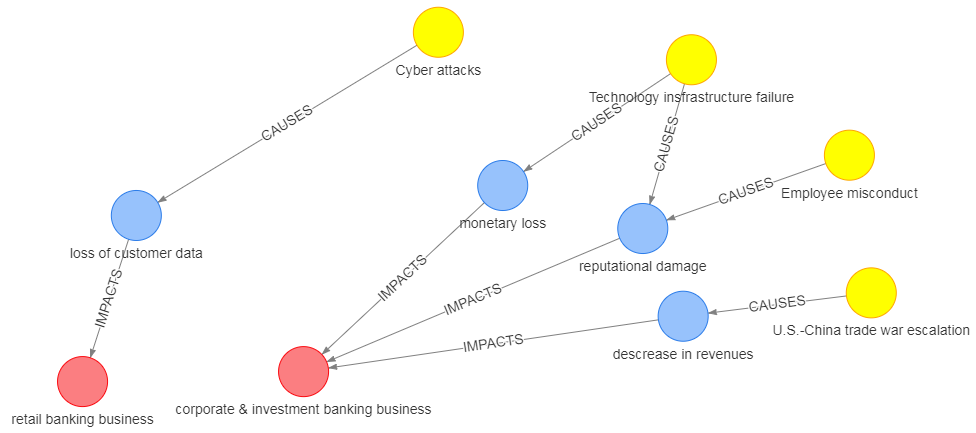}
                \caption{Knowledge graph of the risks in table \ref{table:risks}}
                \label{fig:kg}
            \end{figure}

            We utilize the extracted information to construct a knowledge graph to assist in the risk identification and assessment process. A knowledge graph formally represents semantics by describing entities and their relationships. In our system, the knowledge graph is designed for visualising the risks faced by the institution and for reasoning over data. This is intended to help domain experts understand how risks are related to each other, what are the key triggers of risk facing the institution, etc. The nodes of the graph describe the triggers, outcomes and exposure vessels and the edges describe the relationship between the three categories. In our case, a trigger \textit{causes} a given outcome and the outcome \textit{impacts} a given exposure vessel. Figure \ref{fig:kg} shows the knowledge graph representation for the risks in table \ref{table:risks}.
            
    \subsection{Risk and News Embeddings} \label{sec:sbert}
    
        The news for each risk are retrieved from (1) Google News and (2) The Global Database of Events, Language and Tone (GDELT) \cite{leetaru2013gdelt} based on the trigger identified. This, however, returns a large amount of news with many that are irrelevant to the risk itself. To help filter the news retrieved, we use a neural network model coupled with a cosine similarity metric to identify the top relevant news for each risk. The model used is based on a bidirectional encoder representation from transformers (BERT) neural network \cite{devlin2018bert} which is typically used for predicting masked works in a sentence. Our system uses Sentence-BERT, an extension of the original model used to compute contextual sentence embeddings \cite{reimers2019sentence}. 

\section{Usage \& Conclusion} \label{sec:conclusion}
    We presented a system for institutional risk identification using knowledge graphs and automated news profiling. The system discussed was tested on a set of 1,250 risks faced by our institution for the fourth quarter of 2019. The results were vetted by our business partners who provided strong positive feedback on the output. The model achieved an accuracy of 96.6\% in identifying relevant news on a set of 1,132 news headlines reviewed by domain experts.

\bibliography{aaai22}

\section{Acknowledgments}
    The authors would like to acknowledge Danny Schwartzman, Brian O'toole, Cecilia Tilli, Natraj Raman, Salwa Alamir, Charese Smiley and Andrea Stefanucci from J.P. Morgan for their input and suggestions at various stages of the project.
    
    This paper was prepared for informational purposes by the Artificial Intelligence Research group of JPMorgan Chase \& Co and its affiliates (``J.P. Morgan''), and is not a product of the Research Department of J.P. Morgan. J.P. Morgan makes no representation and warranty whatsoever and disclaims all liability, for the completeness, accuracy or reliability of the information contained herein. This document is not intended as investment research or investment advice, or a recommendation, offer or solicitation for the purchase or sale of any security, financial instrument, financial product or service, or to be used in any way for evaluating the merits of participating in any transaction, and shall not constitute a solicitation under any jurisdiction or to any person, if such solicitation under such jurisdiction or to such person would be unlawful.
\end{document}